\documentclass[lettersize,journal]{IEEEtran}
\usepackage{amsmath,amsfonts}
\usepackage{algorithmic}
\usepackage{algorithm}
\usepackage{array}
\usepackage[caption=false,font=normalsize,labelfont=sf,textfont=sf]{subfig}
\usepackage{textcomp}
\usepackage{stfloats}
\usepackage{url}
\usepackage{verbatim}
\usepackage{graphicx}
\usepackage{cite}
\usepackage{multirow}
\usepackage{bm}
\usepackage{xcolor}
\usepackage{multirow,tabularx}
\usepackage{hhline}
\newcolumntype{C}{>{\centering\arraybackslash}m{1.5cm}}
\newcolumntype{M}{>{\centering\arraybackslash}X}
\newcommand{\vsBf}{\vspace{-0.15in}}
\newcommand{\vsBfSub}{\vspace{-0.15in}}

\usepackage{scalerel}
\usepackage{tikz}
\usetikzlibrary{svg.path}
\definecolor{orcidlogocol}{HTML}{A6CE39}
\tikzset{
  orcidlogo/.pic={
    \fill[orcidlogocol] svg{M256,128c0,70.7-57.3,128-128,128C57.3,256,0,198.7,0,128C0,57.3,57.3,0,128,0C198.7,0,256,57.3,256,128z};
    \fill[white] svg{M86.3,186.2H70.9V79.1h15.4v48.4V186.2z}
                 svg{M108.9,79.1h41.6c39.6,0,57,28.3,57,53.6c0,27.5-21.5,53.6-56.8,53.6h-41.8V79.1z M124.3,172.4h24.5c34.9,0,42.9-26.5,42.9-39.7c0-21.5-13.7-39.7-43.7-39.7h-23.7V172.4z}
                 svg{M88.7,56.8c0,5.5-4.5,10.1-10.1,10.1c-5.6,0-10.1-4.6-10.1-10.1c0-5.6,4.5-10.1,10.1-10.1C84.2,46.7,88.7,51.3,88.7,56.8z};
  }
}

\newcommand\orcidicon[1]{\href{https://orcid.org/#1}{\mbox{\scalerel*{
\begin{tikzpicture}[yscale=-1,transform shape]
\pic{orcidlogo};
\end{tikzpicture}
}{|}}}}

\usepackage[hidelinks]{hyperref} 

\begin{document}

\title{Learning to Evaluate the Artness of AI-generated Images}

\author{
    Junyu~Chen$^{\orcidicon{0000-0003-0145-6045}}$\,,
    Jie~An$^{\orcidicon{0000-0002-1402-8288}}$\,,
    Hanjia~Lyu$^{\orcidicon{0000-0002-3876-0094}}$\,,
    Christopher~Kanan$^{\orcidicon{0000-0002-6412-995X}}$\,,~\IEEEmembership{Senior Member, IEEE}\,,
    and Jiebo~Luo$^{\orcidicon{0000-0002-4516-9729}}$\,,~\IEEEmembership{Fellow,~IEEE}
    \\

\vspace{-6mm}
\thanks{\scriptsize Manuscript received 27 February 2024; revised 20 May 2024; accepted 23 May 2024; data of current version 1 June 2024. The Associate Editor coordinating the review of this manuscript and approving it for publication was Jianlong Fu. {\em(Corresponding author: Jiebo Luo.)}

The authors are with the Department of Computer Science, University of Rochester, Rochester, NY 14627 USA (e-mail: jchen175@ur.rochester.edu, jan6@cs.rochester.edu, hlyu5@ur.rochester.edu, ckanan@cs.rochester.edu, jluo@cs.rochester.edu).

The project page is \url{https://github.com/jchen175/ArtScore}

Digital Object Identifier 10.1109/TMM.2024.3410672}
}

\markboth{IEEE TRANSACTIONS ON MULTIMEDIA,~Vol.~X, 2024}%
{Chen \MakeLowercase{\textit{et al.}}: Learning to Evaluate the Artness of AI-generated Images}

\maketitle

\begin{abstract}
Assessing the artness of AI-generated images continues to be a challenge within the realm of image generation. Most existing metrics cannot be used to perform instance-level and reference-free artness evaluation. This paper presents ArtScore, a metric designed to evaluate the degree to which an image resembles authentic artworks by artists (or conversely photographs), thereby offering a novel approach to artness assessment. We first blend pre-trained models for photo and artwork generation, resulting in a series of mixed models. Subsequently, we utilize these mixed models to generate images exhibiting varying degrees of artness with pseudo-annotations. Each photorealistic image has a corresponding artistic counterpart and a series of interpolated images that range from realistic to artistic. This dataset is then employed to train a neural network that learns to estimate quantized artness levels of arbitrary images. Extensive experiments reveal that the artness levels predicted by ArtScore \textbf{align more closely with human artistic evaluation than existing evaluation metrics}, such as Gram loss and ArtFID.
\end{abstract}


\begin{figure*}[t]
  \includegraphics[width=\textwidth]{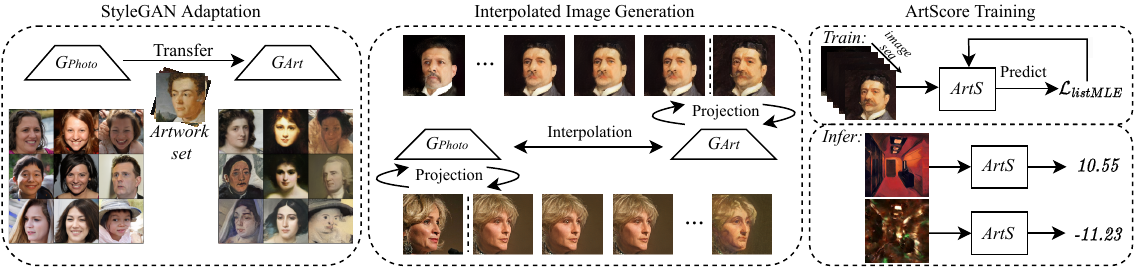}
 \vspace{-0.3in}
  \caption{The proposed framework consists of three steps: 1) StyleGAN transfer, 2) interpolated image generation, and 3) ArtScore training. The first two steps generate a large dataset containing images with varying levels of artness. The third step learns to distinguish different levels of artness from the developed dataset. }
   \vspace{-0.2in}
  \label{fig:framework}
\end{figure*}

\vsBf
\section{Introduction}

\IEEEPARstart{I}{mage}  generation is receiving increasing attention in the field of computer vision. Neural style transfer (NST)~\cite{gatys2015neural,huang2017arbitrary,an2019ultrafast,huang2022quantart}, generative adversarial networks (GANs)~\cite{goodfellow2014generative,elgammal2017can,StyleGAN2,cyclegan}, and diffusion models~\cite{diffusion,dalle2,stable} are three common frameworks that generate images. Significant advances have been made in improving the quality of image generation. However, how to automatically assess the artness of AI-generated images has not been well-explored.
 
Metrics such as Structural Similarity Index Measure (SSIM) \cite{ssim}, Inception Score (IS)~\cite{is}, Fréchet Inception Distance (FID)~\cite{fid}, Content Loss~\cite{gatys2015neural}, ArtFID~\cite{wright2022artfid}, and Gram Loss~\cite{gatys2015neural} are widely used by image generation algorithms to measure the quality of their results. However, \textbf{none} of the existing metrics can be used to perform \textbf{instance-level and reference-free} artness evaluation. FID and IS are usually used to assess the overall quality of the AI-generated images, which are not specially designed for artness evaluation and thus are inaccurate in artness measurement. SSIM and Content Loss are usually used to measure how much content semantics can be preserved in photo-to-art translation, which cannot be used to measure the artness similarity when the contents of two images are different. Gram Loss is used to calculate the artistic distance between the generated image and the real image that is used as the reference when performing style transfer, which has a limited application of image style transfer where the target real artwork is required to perform the artness evaluation. Similar to FID~\cite{fid}, ArtFID can evaluate the distance between the distributions of the generated and real artworks, which may be inaccurate when the number of generated images or the compared real artwork dataset is limited, making instance-level artness evaluation difficult. 

To overcome the lack of automatic artness evaluation metric, subjective evaluation based on a carefully designed user study is widely adopted by existing works~\cite{chandran2021adaptive,an2021artflow,kolkin2019style,liu2021adaattn,huang2022quantart,xu2021drb}. However, human evaluation could be \textbf{expensive} in terms of both time and effort, \textbf{difficult} to reproduce, and might be \textbf{biased} if the number of the collected user responses are not large enough. Therefore, it is desired to have an objective metric that can automatically measure the artness of the algorithm-generated images and is well \textit{aligned} with the subjective human evaluation.
A proper evaluation metric is of critical importance for image generation algorithms. First, it enables fair and qualitative comparison between different algorithms, thus can encourage competition and drive the development of better algorithms. Second, such a metric could be directly used to guide the image generation process to make the generated image have better quality and fidelity.
  
To that end, we propose a new metric to evaluate the artness level of AI-generated images, which is named ArtScore. ArtScore is implemented as a neural network, which allows {\it both} instance-level and reference-free artness evaluation.
To develop the ArtScore model, we first build an interpolated artness dataset and then train a neural network on the newly built dataset to regress the pseudo-annotated artness levels. The whole process can be divided into three steps: StyleGAN Adaptation, Interpolated Image Generation, and ArtScore Training (Fig.~\ref{fig:framework}).
Annotating the artness score of images by humans is very \textbf{expensive} and requires \textbf{extensive expertise}. The first two steps aim to generate a pseudo-annotated dataset to avoid the expensive and time-consuming human annotation~\cite{hu2023promptcap}, where images in our collected dataset exhibit varying levels of artness, with each photorealistic image possessing a corresponding artistic counterpart and a range of interpolated images spanning from photorealistic to artistic.

Starting from a pre-trained StyleGAN2~\cite{StyleGAN2} model with photorealistic style, we first train a model with fine-art painting style through transfer learning, where we use FreezeG~\cite{freezeg} and few-shot adaptation~\cite{ojha2021few} as two transfer learning protocols. FreezeG freezes the early layers of the generator and tunes the later convolutional modules, which empirically yields good results for corresponding domains. The few-shot method can produce different StyleGAN models with varied artistic styles, leading to more diverse artistic images.
For the transfer learning with the few-shot setting, examples to guide the model transfer are representative paintings from different artistic styles and are acquired by artistic style clustering.

Next, we generate a pseudo-annotated artness dataset with varied levels of artness. Based on the pre-trained StyleGAN models with photorealistic style and the adapted model with artistic style, we linearly interpolate~\cite{wang2018esrgan} photorealistic and artistic models to get StyleGAN models that can generate images with varying levels of artness. 
By tuning the interpolation weight between the base model and the adapted model, we can control the level of artness of the mixed model, allowing us to generate images with same structures but different levels of artness using the same latent input. We project the photorealistic images from FFHQ~\cite{karras2019style}, LSUN horse~\cite{yu2015lsun}, and LHQ~\cite{skorokhodov2021aligning} datasets into the latent space of the pre-trained StyleGAN models~\cite{StyleGAN2} and generate images with varying artness levels by using the interpolated StyleGAN models from photo to art.

We train a neural network (ArtScore) to predict how close an image is to real artwork, which can vary based on subjective interpretations and criteria.  
However, it is challenging to define a fixed absolute artness ground truth in model training. Therefore, we train the model with a \textbf{learn-to-rank} objective~\cite{xia2008listwise} instead of regressing the absolute artness level of the interpolated images, where the model is trained to rank the \textbf{relative artness} of a given image sequence. The learn-to-rank loss term avoids the need for fixed absolute artness ground truth but can still learn effective artness scores. To further improve the model's robustness during inference, we introduce a novel augmentation strategy. During training, an image in a sequence is randomly replaced with an image from another sequence with the same relative artness rank but different content and style. This scenario helps the model to compare the artness of less related images. 
Our ablation study demonstrates that using learn-to-rank loss terms produces better results than regressing fixed artness levels, and the cross-sequence random swap augmentation further enhances the model performance.

We evaluate the effectiveness of our proposed metric at both the instance level and algorithm level, demonstrating that the artness level measured by ArtScore is better \textbf{aligned with the human evaluation}.
Our experiment also shows that using ArtScore together with other image quality assessment metrics can produce a more accurate image artness ranking, which \textbf{cannot be achieved by other metrics} without ArtScore. Additionally, we conduct an ablation study to evaluate the effectiveness and necessity of our framework design.

In summary, our contributions are three folds:
\begin{itemize}
\item We develop a large dataset that facilitates effective learning of the proposed artness metric, without relying on any human annotation which can be expensive and require expertise.
\item We propose ArtScore, which can quantitatively evaluate how well a generated image resembles fine-art paintings. It is a neural network trained with an effective learn-to-rank objective. To the best of our knowledge, ArtScore is the first method that can achieve reference-free instance-level artness evaluation.
\item We empirically demonstrate that ArtScore aligns better with human evaluation than existing metrics and can be used to better evaluate neural style transfer algorithms quantitatively when combined with other existing metrics.

\end{itemize}

\vsBf
\section{Related Work}
\subsection{Image Generation Frameworks}
Image generation has gained significant interest in recent years. Neural Style Transfer (NST) is a prominent method that transfers artistic style from a style image to a content image. \cite{gatys2015neural} introduces an iterative algorithm to optimize gram loss and content loss. ~\cite{johnson2016perceptual} and~\cite{ulyanov2016texture} introduce feed-forward networks for specific style approximation. Subsequent techniques allow multiple styles per model~\cite{huang2017real,gong2018neural}, and universal style transfer methods~\cite{huang2017arbitrary,sheng2018avatar,an2019ultrafast,an2020real}. Generative Adversarial Networks (GANs) represent another popular approach, which involves two competing neural networks - a generator and a discriminator. Various advanced and domain-specific GAN variations have been utilized for image generation, such as CAN~\cite{elgammal2017can}, StyleGAN~\cite{StyleGAN}, and CycleGAN~\cite{cyclegan}, which achieves impressive results in generating convincing fake artwork images. The diffusion model~\cite{diffusion} is a recently proposed framework with notable potential. It involves creating a series of progressively noisier versions of the original images and learning to denoise them iteratively. DALL-E 2~\cite{dalle2}, stable diffusion~\cite{stable}, MultiDiffusion~\cite{bar2023multidiffusion}, and RAPHAEL~\cite{xue2024raphael} are text-to-image generation methods based on the diffusion model.
To train the ArtScore model, we develop a dataset with StyleGAN2 models. In our experiments, to evaluate the trained model, we use various NST methods and StyleGAN to generate images for evaluation.

\vsBfSub
\subsection{Image Generation Evaluation}
Several metrics have been proposed to evaluate the generated images based on specific dimensions. Inception Score (IS)~\cite{is} and Fréchet Inception Distance (FID)~\cite{fid} are based on the image set feature distribution. IS compares the conditional label distribution to the marginal distribution of the generated images to reflect quality and diversity. FID measures the distance between the deep feature distributions of the generated and real image sets.
Structural Similarity Index Measure (SSIM)~\cite{ssim} and Learned Perceptual Image Patch Similarity (LPIPS)~\cite{lpips} are based on the comparison of two images. SSIM measures the similarity between the generated and real images on the pixel level, considering luminance, contrast, and structure. LPIPS calculates the average l2 distance between the patch-level feature representations.
Content Loss~\cite{gatys2015neural} and Gram Loss~\cite{gatys2015neural} are commonly used metrics in NST research~\cite{an2021artflow,deng2022stytr2,huang2017arbitrary,hong2021domain}. Content Loss measures the preservation of content between the transferred image and the content reference based on MSE of their feature maps. Gram Loss measures the style matching between the stylized image and the style reference, based on the MSE of the gram matrices of their feature maps.
Deception Rate~\cite{sanakoyeu2018style} and ArtFID~\cite{wright2022artfid} utilize pre-trained stylistic classifiers trained on art painting datasets to improve style representation. Deception Rate measures the proportion of stylized images that an artist classification network assigns to the artist whose artwork was used for stylization. ArtFID incorporates both content and style by computing the LPIPS distance between the content reference and the stylized image for content preservation, and the FID between the style reference image set and the stylized image set for style matching (ArtFID\_Style). The resulting evaluation metric is computed as $ArtFID=(1+LPIPS)\cdot(1+ArtFID\_Style)$. 
To evaluate the effectiveness of our proposed ArtScore, we compare it against commonly used metrics in artness evaluation.

\vsBfSub
\subsection{StyleGAN Inversion and Transfer}  
Latent space projection is a widely used technique in StyleGAN~\cite{StyleGAN} to attribute a given image. It involves finding a matching latent code for a given image with the generator; so that when feeding the latent code back into the generator, it can produce an image that closely matches the given target image. This technique is useful to control attributes of the generated image in image editing~\cite{huh2020transforming}. \cite{StyleGAN2} explored an optimization-based method for latent space projection, where the generator is fixed while the latent code is optimized to minimize the LPIPS between the generated image and the given target.

Transfer learning has been employed to adapt pre-trained StyleGAN models to new styles~\cite{ojha2021few,gal2022StyleGAN,xiao2022few}, enabling the training of high-quality generators with limited examples. The transferred generator closely relates to the source generator: by linearly interpolating between their weights, network interpolation allows for an approximate interpolation between the two learned styles~\cite{wang2018esrgan}. \cite{pinkney2020resolution} demonstrates that by fusing different layers, one can control the structural and stylized visual effects of the fused generator to resemble either the source or the transferred model.

We use StyleGAN transfer and interpolation to derive generators with varied artness levels, and use StyleGAN inversion to retrieve latent codes for real images as the input for these generators to synthesize the training dataset.

\vsBfSub
\subsection{Learning to Rank}
Learning to rank (LTR)~\cite{cao2007learning} aims to sort a list of objects based on a scoring model. It is commonly used in various applications such as recommendation systems and information retrieval. During the training process, the scoring model learns to predict scores for a list of input objects by minimizing a ranking loss function. LTR methods differ in how they consider objects in the loss function. Compared with pointwise and pairwise approaches, which transfer ranking into regression on a single object or classification on object pairs, the listwise approach such as ListMLE~\cite{xia2008listwise} takes the whole list as input, which better considers the relative position information~\cite{xia2008listwise,qin2010letor} and leads to better experimental results. 
The proposed ArtScore is trained based on the optimization of a listwise LTR objective.

\begin{figure}[t]
	\centering
	\includegraphics[width=0.4\textwidth]{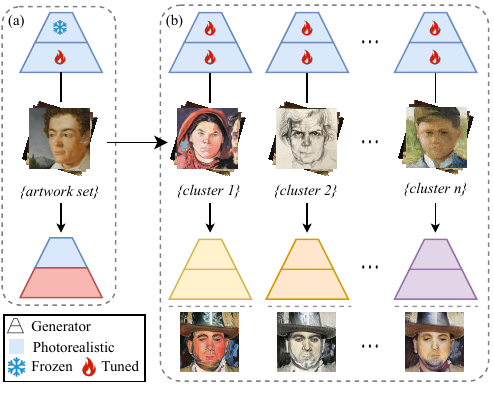}
  \vspace{-0.2in}
	\caption{We use (a) FreezeG and (b) a few-shot adaptation method to adapt the photorealistic StyleGAN to artistic styles. In the FreezeG setting (a), we fine-tune the photorealistic model using the entire art painting dataset while freezing the early convolution layers. In the few-shot adaptation setting (b), we cluster the art paintings by their styles and sample representative paintings to fine-tune the entire photorealistic model using only a few representative examples.}
  \vspace{-0.2in}
	\label{fig:freeze_few}
\end{figure}

\vsBf
\section{Method}
Fig.~\ref{fig:framework} shows the whole process to compute ArtScore. We first use transfer learning procedures to adapt the pre-trained photorealistic StyleGAN models to the artistic style.
Next, we blend pre-trained StyleGAN models for photo and artwork generation and use the blended model to generate a dataset exhibiting varying levels of artness, where the model interpolation ratios are used as the pseudo annotations of the artness level.
Finally, we train a neural network to measure the artness of images based on a learn-to-rank objective. This section introduces the details of each step.

\vsBfSub
\subsection{StyleGAN Transfer}
We adapt pre-trained StyleGAN2~\cite{StyleGAN2} models from the photorealistic style to the artistic style through transfer learning. Given the same latent code in the latent space of StyleGAN, the original pre-trained model and the transferred model generate photorealistic and artistic images, respectively, where both images have the same content.
We perform the transfer learning on StyleGAN models pre-trained on FFHQ~\cite{karras2019style}, LSUN horse~\cite{yu2015lsun}, and LHQ~\cite{skorokhodov2021aligning} datasets because they are popular genres in art paintings. Paintings corresponding to these domains are collected from the WikiART~\cite{WikiART} dataset and used to transfer the models. We explore two transfer learning methods: FreezeG~\cite{freezeg} and few-shot adaptation~\cite{ojha2021few} (see Fig.~\ref{fig:freeze_few}).

FreezeG freezes early layers of the generator and tunes the last generator blocks, which can preserve the latent code structure intact while transferring the StyleGAN into an artistic style. 
The few-shot method needs a few examples to transfer the photorealistic generator into an artistic style, where we can control the transferred model by using artworks with a certain artistic style as examples. By switching the example images, we adapt the same photorealistic model into different artistic models that produce distinct styles, leading to more diverse artistic images. Representative paintings from different artistic styles are selected via clustering for this purpose. 
Specifically, we adopt the Gram matrix~\cite{gatys2015neural} (stacked into a 1-d vector) as the style representation and use the K-Means algorithm for clustering. Paintings closest to the cluster centers are used as representatives of the cluster to adapt the photorealistic model. We perform the few-shot StyleGAN model transfer on all clusters, resulting in one pre-trained photorealistic StyleGAN model with corresponding artistic models in different artistic styles.

\begin{figure}[t]
	\centering
	\includegraphics[width=0.48\textwidth]{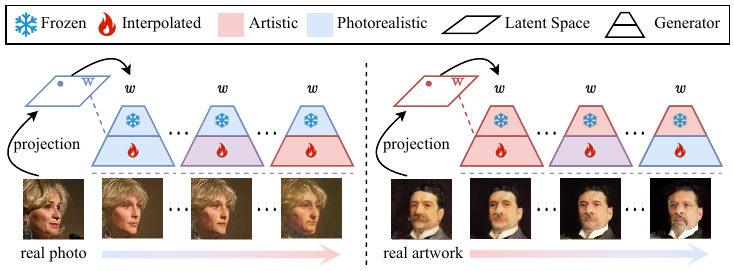}
   \vspace{-0.1in}
	\caption{StyleGAN model interpolation and interpolated image generation. We linearly interpolate the low-level convolution layers of photorealistic and artistic StyleGAN models. Real photos and artworks are projected into the latent space of respective models to retrieve latent codes. These codes are used as the input of the interpolated models to generate sequences of images with varied levels of artness.}
   \vspace{-0.1in}
	\label{fig:interpolate_method}
\end{figure}

\begin{figure}[t]
	\centering
	\includegraphics[width=0.40\textwidth]{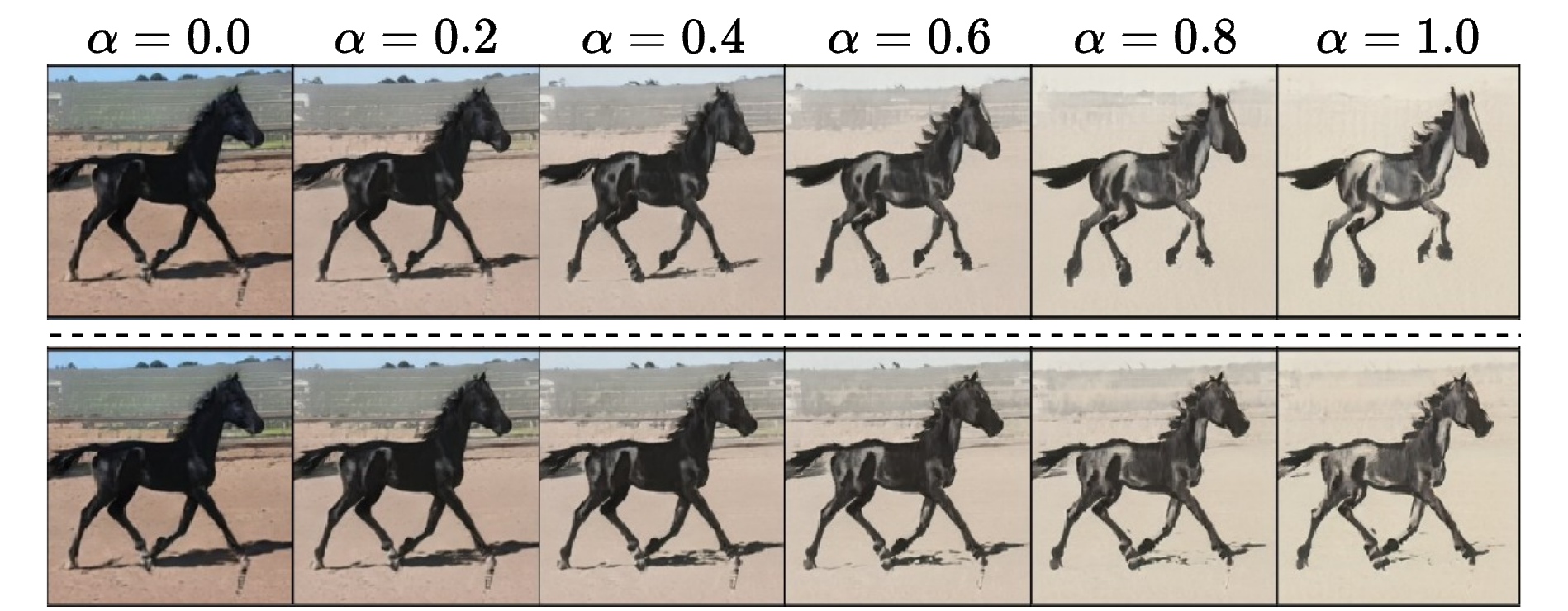}
  \vspace{-0.1in}
	\caption{Effect of fusing all blocks (top) or the last blocks (bottom) of the photorealistic and artistic models. Note that the structure of the horse is better preserved if we only fuse the low-level layers.}
 \vspace{-0.2in}
	\label{fig:fuse}
\end{figure}

\vsBfSub
\subsection{Pseudo-Annotated Dataset Generation}
This subsection introduces how we use the photorealistic and artistic StyleGAN models to synthesize a variety of images, including photorealistic images, artistic images, and intermediate images between the two styles with varying artness levels, as shown in Fig.~\ref{fig:interpolate_method}.
We first interpolate network parameters of the photorealistic and artistic StyleGAN models~\cite{wang2018esrgan} to derive mixed models that generate interpolated images between real and art styles. The parameters of the interpolated model are a weighted average of the photorealistic model and artistic model:
\begin{equation} 
\theta_i^{I}=(1-\alpha)\cdot\theta_i^{P}+\alpha\cdot\theta_i^{A},
\label{eq:interpolation}
\end{equation}
where $\theta_i^{I}$, $\theta_i^{P}$, and $\theta_i^{A}$ are the parameters of the $i^{th}$ convolutional layer of the interpolated, photorealistic, and artistic models, respectively. $\alpha$ is the interpolation weight to control the desired artness level, which also serves as the pseudo-annotation.

Next, we project a set of real paintings and photos to the latent space of the corresponding StyleGAN models in the artistic and photorealistic domains, respectively. We use the StyleGAN inversion algorithm introduced by~\cite{abdal2019image2StyleGAN}. For the artistic StyleGAN models produced by the few-shot transfer learning method, we project the paintings that are used as examples for StyleGAN transfer.

Finally, based on the interpolated StyleGAN models and retrieved latent codes, we use StyleGAN models with varying levels of artness to generate images with pseudo artness labels. The process starts with a real painting or photo that serves as the ground truth of most or least artness, respectively. Next, we generate a series of images from the original input (\textit{i.e.,} real painting or photo) while transitioning smoothly to the other style. 
Given the same latent code, the interpolated StyleGAN models will generate images with the same semantic content but varying artness levels. As shown in Fig.~\ref{fig:fuse}. This is achieved by only fusing the last 6 convolutional layers while keeping the parameters of the previous layers intact.

\begin{table}[tbp]
\caption{Summary of the generated pseudo-annotated artness dataset. We provide the number of sequences containing either real paintings or real photos for each domain. Each sequence consists of 12 images with the same content and different styles that interpolate between the photorealistic and artistic styles.}

\resizebox{\linewidth}{!}{
\begin{tabularx}{0.48 \textwidth}{l|>{\centering\arraybackslash}X>{\centering\arraybackslash}X|>{\centering\arraybackslash}X}\hline
\textbf{Domain} & \textbf{Painting Seq} & \textbf{Photo Seq} & \textbf{Total Seq} \\ \hline
Face & 17,436 & 17,436 & 34,872 \\
Horse & 8,808 & 8,808 & 17,616 \\
Landscape & 45,627 & 45,627 & 91,254 \\ \hline
All & 71,871 & 71,871 & 143,742\\\hline
\end{tabularx}
}
\vspace{-0.2in}
\label{table:dataset summary}
\end{table}
 
\vsBfSub
\subsection{Model Training}
This subsection shows how to train a neural network on the pseudo-annotated dataset to develop the ArtScore model. Directly regressing the pseudo artness label with the corresponding image as the input does not achieve satisfying results. The reason is that different real artworks may exhibit different artness levels. However, in our collected pseudo-annotated dataset, all the real artworks have the same artness level, which may confuse the network in training, leading to compromised accuracy. To this end, we adopt a learn-to-rank scheme to train a neural network to predict the relative ranking of images rather than fixed absolute artness levels. Given an interpolated image sequence ${\mathbf{x}}=\{{x}_{1}, \cdots,{x}_{n}\}$ and the corresponding artness levels ${\mathbf{y}}=\{{y}_{1}, \cdots,{y}_{n}\}$, the objective is to learn a model $f$ that can output artness scores that are close to the ground-truth ranking. To achieve this, we use the ListMLE loss, which is defined as 
\begin{equation}
\mathcal{L}(f; {\mathbf x}, {\mathbf y}) = -\log P({\mathbf y}|{\mathbf x};f),
\end{equation}
where 
\begin{equation}
P({\mathbf y}|{\mathbf x};f)=\prod_{i=1}^{n} \frac{ \exp(f(x_{{\mathbf y}^{-1}(i)}))}{\sum_{k=i}^n \exp(f(x_{{\mathbf y}^{-1}(k)}))}
\end{equation}
Here, ${\mathbf y}^{-1}(i)$ denotes the index of image in the $i$-th position of $\mathbf y$, so $f(x_{{\mathbf y}^{-1}(i)})$ represents the model output score of the $i$-th ranked image in the given sequence. 
To help the model compare the artness between images with different contents and styles, we introduced a random cross-sequence swap augmentation during training, where an image in a sequence (\textit{i.e.,} the image sequence generated by StyleGAN model interpolation) is randomly replaced with an image from another sequence with the same relative artness rank but different content and style. Our experiments find that the learn-to-rank method can achieve better results than the intuitive regression framework, and the random swap augmentation enhances the model's robustness during inference and improves its performance.

\vsBf
\section{Experiment}
An effective evaluation metric should closely align with human judgment~\cite{lpips,wright2022artfid,lin2023videoxum}. To validate the efficacy of our framework, we compare the artness level produced by ArtScore with other metrics on a human-annotated artness dataset. The results demonstrate that the proposed ArtScore aligns better with the human measurements in terms of the artness. Using ArtScore with other image quality assessment metrics together achieves more accurate and comprehensive AI-generated image quality evaluation. 

\vsBfSub
\subsection{Implementation Details}
\noindent \textbf{Dataset.} WikiART~\cite{WikiART} is a large-scale dataset of over 80k unique paintings with detailed metadata on artists, styles, and genres. We collect a set of art paintings including art faces, art horse paintings, and landscape paintings from the WikiART dataset to perform the transfer learning on pre-trained StyleGAN2~\cite{StyleGAN2} models from photorealistic style to artistic style. For the art face domain, we leverage an available and preprocessed dataset~\cite{wikiartface}. To collect art horse paintings, we retrieved $2,000$ paintings from WikiART whose CLIP~\cite{radford2021learning} embeddings had the closest cosine similarity to the CLIP embedding of the prompt ``this is a horse painting'', as well as all paintings with the keyword ``horse'' in the accompanying metadata (\textit{i.e.,} genre, name, description, and tags). We manually exclude false positive horse paintings. For the landscape painting domain, we use all paintings belonging to the ``landscape'' genre in WikiART. In total, we collect $5,812$ art faces, $2,936$ horse paintings, and $15,209$ landscape paintings.

\noindent \textbf{StyleGAN Transfer Learning.} We use two settings to transfer the pre-trained photorealistic StyleGAN2 models to the artistic models. In the first setting, we use FreezeG technique, which fixes the early layers of the generator and fine-tunes the last generator blocks. We determine which layers to fuse through hyperparameter tuning, and our experiments show that tuning only the last $5$ or $6$ blocks or the entire model produces visually pleasing results. For this method, we use a learning rate of $0.0002$, a batch size of $16$, and a total of $10,000$ training iterations, and use all collected paintings belonging to the corresponding domains for adaptation. In the second setting, we use the few-shot adaptation method. First, we cluster the paintings using K-Means based on their artistic styles represented by the Gram matrices. We set the number of clusters to $20$, which approximates the number of unique artistic styles in the dataset. For each cluster, we select 10 paintings that are closest to the cluster center as representatives and use them to adapt the photorealistic model. We perform the few-shot StyleGAN model transfer on all $20$ clusters using a learning rate of $0.0002$, a batch size of $4$, and a total of $5,000$ training iterations. We keep all other hyperparameters the same as the official implementations.

\noindent \textbf{Interpolated Image Generation.}  We use the interpolation parameter $\alpha$ to control the level of artness in the generated images, which ranges from 0.0 to 1.0 in increments of 0.1. This results in 12 images per sequence, including one real photo or one painting, and 11 interpolated images with varying degrees of artness. Table~\ref{table:dataset summary} shows the statistics of the developed datasets.

\noindent \textbf{ArtScore Training.} We use a ResNet50 pre-trained on ImageNet as the backbone of the proposed ArtScore, with the fully connected layer replaced by a three-layer MLP with a hidden layer of size $768$ and an output layer of size $1$. We use the AdamW optimizer~\cite{loshchilov2017decoupled} with a learning rate of $1\mathrm{e}{-4}$ to optimize the ListMLE loss~\cite{xia2008listwise}. The batch size is set to $16$ sequences (\textit{i.e.,} $192$ images). The dropout rate is set to $0.5$ to prevent over-fitting. Apart from the random swap augmentation, commonly used augmentations including random flipping and rotation are applied to the training images. 
The dataset is randomly split into training, validation, and test sets with a ratio of $0.8$, $0.1$, and $0.1$. We train the model for $10$ epochs and save the model parameters when the model reaches the lowest loss on the validation set as the final evaluator, which reaches an NDCG score of $0.965$ on the test set.

\vsBfSub
\subsection{Quantitative Consistency with Human Evaluation}
\cite{wright2022artfid} systematically compared the effects of the photo-to-art style transfer of 13 different NST algorithms through a large-scale user study. 
We adopt the user study score presented in their paper as the oracle measurement of the quality of these algorithms in generating artwork. We then evaluate how our proposed ArtScore and other existing metrics align with the human evaluation by measuring Spearman’s rank correlation coefficient~\cite{spearman1961proof}. Following the experimental setup in \cite{wright2022artfid}, we generate 50,000 $512\times512$ style transfer images for 12 style transfer methods from the literature~\cite{luo2022consistent,chen2017coherent,liu2021adaattn,svoboda2020two,deng2020arbitrary,park2019arbitrary,sheng2018avatar,huang2017arbitrary,huo2021manifold,li2019learning,gatys2015neural,li2017universal} using the same pairs of content and style references. We exclude one method~\cite{chen2022towards} due to insufficient implementation details.
Content images are sampled from the Places365 dataset~\cite{zhou2017places} and the COCO dataset~\cite{lin2014microsoft}. Style images are sampled from the WikiART~\cite{WikiART} dataset. 
We compute three different sets of metrics: (1) content preservation metrics including LPIPS, Content Loss, SSIM, and L2 distance, calculated using the stylized images and the content references; (2) style matching metrics including ArtFID\_Style (\textit{i.e.,} the style measure used in the ArtFID metric) and Gram Loss, calculated using the stylized images and the style references; (3) artness metric based on the proposed ArtScore, which \textit{does not require any references}. To obtain an aggregated ArtScore metric for an algorithm, we first calculate the artness score for each stylized image using the trained ArtScore, then pass the scores through a sigmoid function and calculate their average. Therefore, the final ArtScore for an algorithm is normalized to a range between 0 and 1, with a higher score indicating that an algorithm is more capable of generating images that resemble fine-art paintings.

\begin{table}[t]
\small
\centering

\caption{The correlation between the ranking induced by the user study conducted in~\cite{wright2022artfid} and the ranking induced by ArtScore and other metrics based on the Rank aggregation method given by Eq.~\ref{equ:rank}. For the ArtFID metric, we report the result in~\cite{wright2022artfid}(paper), and the re-implemented result using the stylized images generated in our experiments (rep). The highest correlation is highlighted in bold, and the second highest is underlined. }
\label{table:correlation result}

\setlength\tabcolsep{12pt}
\resizebox{\linewidth}{!}{
\begin{tabular}{c|l|lc}
\hline
\textbf{Aspects} &  \multicolumn{1}{c|}{\textbf{Metrics}} & \textbf{Spearman’s \bm{$\rho$} ↑} & \textbf{\textit{p}-value ↓}\\\hline
Artness & ArtScore & 0.6364 & 0.0261\\\hline
\multirow{3}{*}{Style} & Gram Loss & -0.3286 & 0.2969 \\\cline{2-4}
 & ArtFID\_Style & 0.6504 & 0.0220 \\
 & \hspace{3mm}+ArtScore & 0.7549 (\textcolor{red}{+0.1045}) & 0.0045 \\\hline
\multirow{8}{*}{Content} & LPIPS & 0.4126 & 0.1826 \\
 & \hspace{3mm}+ArtScore & 0.7138  (\textcolor{red}{+0.3012}) & 0.0091 \\\cline{2-4}
 &  Content Loss (CL) & 0.4755 & 0.1182 \\
 & \hspace{3mm}+ArtScore & 0.5835  (\textcolor{red}{+0.1080}) & 0.0464 \\\cline{2-4}
 & SSIM & 0.5315 & 0.0754 \\
 & \hspace{3mm}+ArtScore & 0.5965 (\textcolor{red}{+0.0650})& 0.0406 \\\cline{2-4}
 & L2 & 0.2837 & 0.3715 \\
 & \hspace{3mm}+ArtScore & 0.5298 (\textcolor{red}{+0.2461}) & 0.0764 \\\hhline{====} 
 \multirow{11}{*}{Style+Content} & ArtFID$\infty$ (paper) & 0.9301* & 0.0000*\\\cline{2-4}
 & ArtFID$\infty$ & 0.7063 & 0.0102 \\
 & \hspace{3mm}+ArtScore & 0.7790 (\textcolor{red}{+0.0727}) & 0.0028 \\\cline{2-4}
 & ArtFID\_Style+LPIPS & 0.6714 & 0.0168 \\
 & \hspace{3mm}+ArtScore & 0.7552 (\textcolor{red}{+0.0489}) & 0.0045 \\\cline{2-4}
 & ArtFID\_Style+CL & 0.7874 & 0.0024 \\
 & \hspace{3mm}+ArtScore & \textbf{0.8196}  (\textcolor{red}{+0.0322}) & \textbf{0.0011} \\\cline{2-4}
 & ArtFID\_Style+SSIM & \underline{0.8120} & \underline{0.0013} \\
 & \hspace{3mm}+ArtScore & 0.7719 (\textcolor{black}{-0.0401}) & 0.0033 \\\cline{2-4}
 & ArtFID\_Style+L2 & 0.6843 & 0.0141 \\
 & \hspace{3mm}+ArtScore & 0.7356 (\textcolor{red}{+0.0513}) & 0.0064 \\\hline
\end{tabular}
}
\vspace{-0.2in}
\end{table}

\begin{figure}[t]
	\centering
	\includegraphics[width=0.46\textwidth]{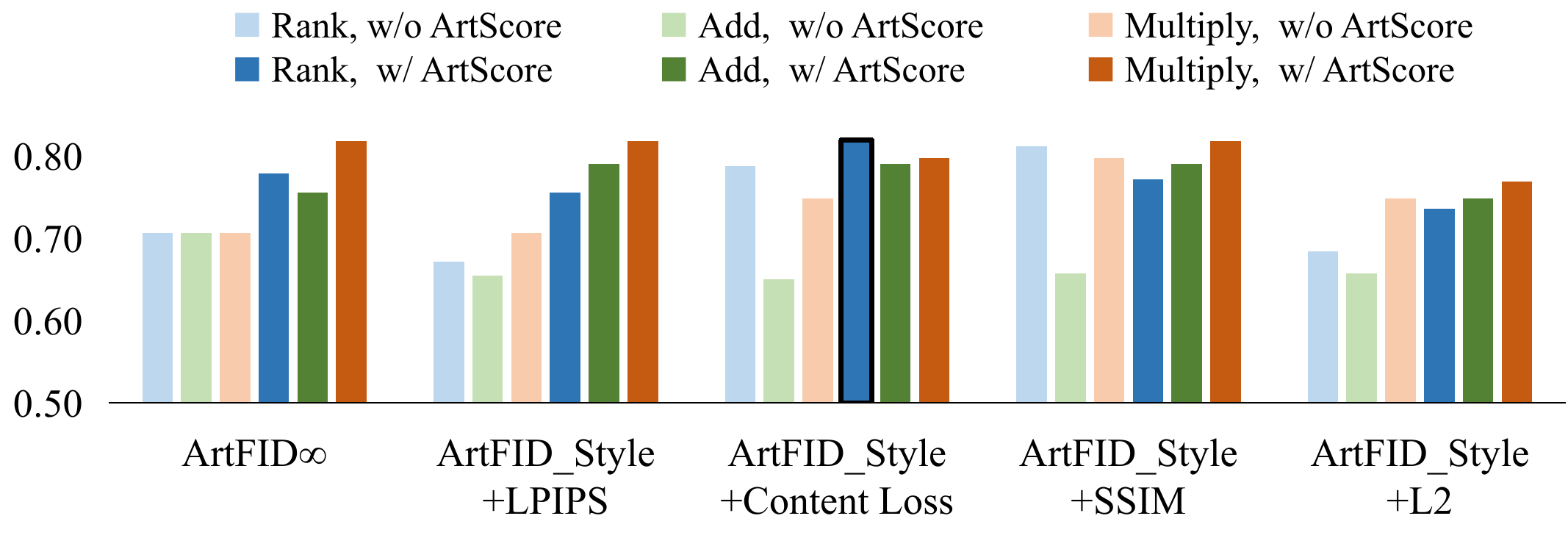}
   \vspace{-0.1in}
	\caption{The correlation between the user study rankings from~\cite{wright2022artfid} and rankings induced by content preservation and style matching metrics, both with and without our proposed ArtScore metric. We tested three aggregation methods denoted by Eq.~\ref{equ:rank}, Eq.~\ref{equ:add}, and Eq.~\ref{equ:mul}. Lighter shades represent results without ArtScore, while darker shades represent results with ArtScore. In general, incorporating ArtScore consistently improves the accuracy of image artness ranking across all aggregation methods.}
   \vspace{-0.2in}
	\label{fig:quantitative}
\end{figure}

To measure the alignment of metrics and their combinations with human evaluation scores, we calculate Spearman’s $\rho$ correlation coefficient, and the results are presented in Table~\ref{table:correlation result} and Fig.~\ref{fig:quantitative}. For a single metric, we can calculate the correlation directly, while for multiple metrics, we explore three methods for combining them.

\noindent\textbf{Rank.} We denote $M_i$ as the $i$-th metric and $Alg_j$ as the $j$-th algorithm. We calculate the rank of each algorithm $Alg_j$ based on each metric $M_i$, which is denoted as $R_{ij}$.
The overall rank $R_j$ of algorithm $Alg_j$ is then defined as:
\vspace{-0.08in}
\begin{equation}
\label{equ:rank}
    R_j = \sum_{i=1}^{n} R_{ij}
    \vspace{-0.08in}
\end{equation}
where $n$ is the total number of metrics used. $R_j$ is then used as the aggregated score for calculating the correlation.

\noindent\textbf{Add.} Here, we first normalize the metric values between 0 and 1 and make them smaller-the-better form, denoted by ${M_{norm}}$. We denote ${M_{norm}}_{ij}$ as the $j$-th algorithm's value on $i$-th normalized metric.
We then use
\vspace{-0.08in}
\begin{equation}
\label{equ:add}
    A_j = \sum_{i=1}^{n} {M_{norm}}_{ij}
\vspace{-0.08in}
\end{equation}
to combine the normalized metrics to obtain an aggregated score.

\noindent\textbf{Multiply.} Here we also use normalized metric values and use
\vspace{-0.08in}
\begin{equation}
\label{equ:mul}
Mul_j = \prod_{i=1}^{n}(1+{M_{norm}}_{ij})
\vspace{-0.08in}
\end{equation}
to combine the normalized metrics. This is the formula used in ArtFID to combine LPIPS and ArtFID\_Style metrics, ensuring a fair comparison.

As Table~\ref{table:correlation result} shows, when a single metric is used to evaluate style transfer algorithms, ArtScore and ArtFID\_Style exhibit a significant correlation with human judgment and perform better than content-related metrics. This could be because both metrics are based on models trained on large-scale art painting datasets and have abundant priors about artistic styles.
As shown in Table~\ref{table:correlation result} and Fig.~\ref{fig:quantitative}, when combined with other metrics, ArtScore is proved to be effective in improving the correlation with human judgments under different aggregation settings. Specifically, combining the rank induced by ArtScore with LPIPS yields a higher correlation coefficient than the strong baseline ArtFID, which consists of LPIPS and ArtFID\_Style. Moreover, combining ArtScore with ArtFID further improves the correlation. These results show that ArtScore can accurately and quantitatively evaluate the quality of images generated by style transfer methods.

Interestingly, we observe that the Gram Loss metric exhibits a negative correlation with human evaluation scores, and incorporating it into combinations with other metrics would reduce the correlation. This suggests that Gram Loss may not be aligned well with human artistic evaluation. A similar pattern is shown in previous research~\cite{an2021artflow,deng2022stytr2,hong2021domain}, where methods with lower Gram Loss did not necessarily result in better preference or style preservation in human evaluation studies.

\vsBfSub
\subsection{User Study}
We conduct a user study to evaluate the consistency between our proposed ArtScore and existing metrics with human artistic evaluation at the instance level. Participants were presented with pairs of images and asked to select which one more closely resembled fine-art paintings. The comparisons included three different scenarios: pairs of images generated with the same, different, or no style-content references.  We use the same 12 style transfer methods as in the algorithm-level experiments to produce art images with the same or different references and utilize a pre-trained StyleGAN3~\cite{ArtStyleGAN} to synthesize art images unconditionally.  
\begin{table}[t]
\centering
\caption{We evaluate the alignment of each metric with human artistic judgments by using the rate at which its artness comparison results match those of humans. The proposed ArtScore achieves the best result. We use the McNemar test to examine the differences between the ArtScore and other metrics in artistic evaluation.}
\vspace{-0.1in}
\begin{tabularx}{0.48 \textwidth}{X|X|ccc}
\hline
\multicolumn{1}{c|}{\textbf{Comparision}} & \multicolumn{1}{c|}{\textbf{Metrics}} & \textbf{Accuracy} & \bm{$\chi^2$} & \bm{$p$}\textbf{-value} \\ \hline
\multirow{6}{*}{\begin{tabular}[c]{@{}c@{}}NST with \\ same style \\ and content \\ references\end{tabular}} & ArtScore & \textbf{0.6433} & - & - \\ \cline{2-5} 
 & Gram & 0.4503 & 21.1250 & 0.0000 \\
 & LPIPS & 0.5848 & 2.5423 & 0.1108 \\
 & Content & 0.5819 & 2.5478 & 0.1105 \\
 & SSIM & 0.6053 & 1.0992 & 0.2944 \\
 & L2 & 0.5731 & 3.7786 & 0.0519 \\ \hline
\multirow{6}{*}{\begin{tabular}[c]{@{}c@{}}NST with \\ different style \\ and content \\ references\end{tabular}} & ArtScore & \textbf{0.5639} & - & - \\\cline{2-5} 
 & Gram & 0.4511 & 5.9225 & 0.0149 \\
 & LPIPS & 0.5301 & 0.5378 & 0.4633 \\
 & Content & 0.4361 & 8.6429 & 0.0033 \\
 & SSIM & 0.4699 & 4.8403 & 0.0278 \\
 & L2 & 0.5226 & 0.8547 & 0.3552 \\ \hline
StyleGAN3 & ArtScore & \textbf{0.6426} & - & - \\ \hline
\end{tabularx}
\vspace{-0.2in}
\label{table:evaluation result}
\end{table}

To ensure the reliability of the data, an attention check question is included in the user study. Each participant is given a total of 46 questions, consisting of 15 pairs of images for each of the three types of comparisons (same, different, and no style-content references), as well as the attention check question. 
We remove responses that do not pass the attention check.
We also exclude pairs of images where an equal number of participants preferred one image over the other to avoid any controversial responses.

\begin{figure*}[t]

	\centering
	\includegraphics[width=0.95\textwidth]{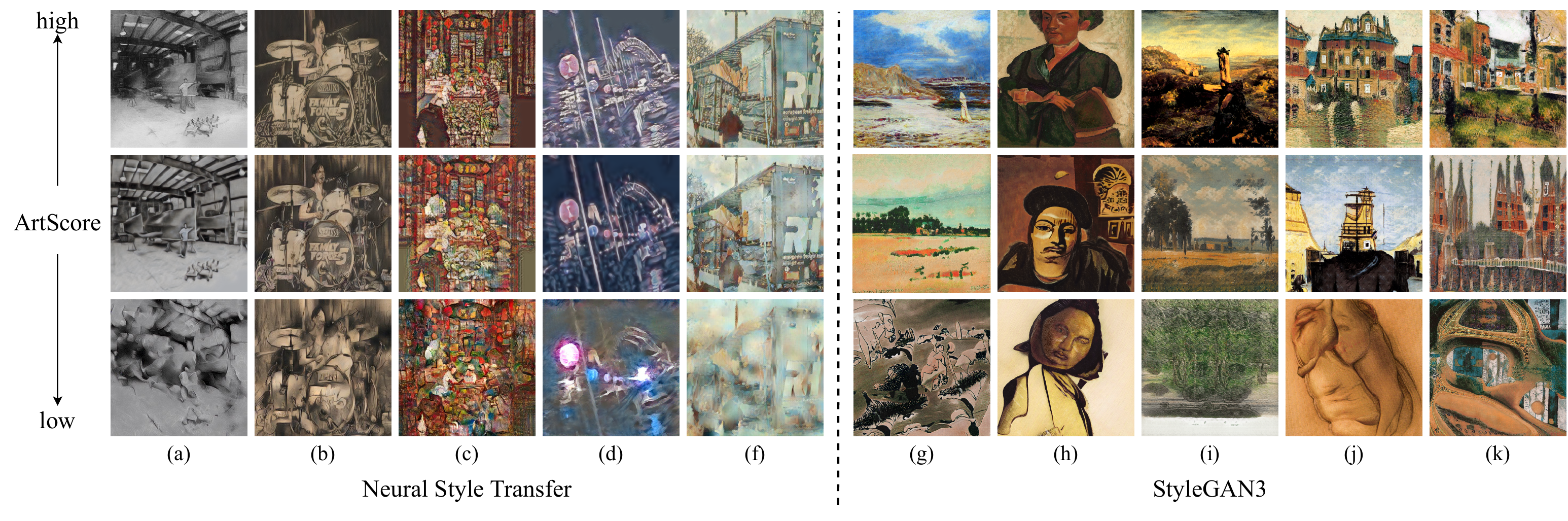}
 \vspace{-0.2in}
	\caption{Qualitative comparisons on images that are conditionally generated (with neural style transfer algorithms) and unconditionally generated (with StyleGAN3).  A primary difference between images with low and high ArtScore values is that high-score images tend to resemble fine-art paintings more closely. This can be observed in more structured composition, more delicate brushstrokes, and fewer artifacts.}
	\label{fig:qualitative}
 \vspace{-0.2in}
\end{figure*}

To evaluate the consistency of each metric with human artistic evaluation, we use the artness comparison results from human evaluation as the reference standard. We then calculate an accuracy score for each metric based on its induced comparison results. We also conduct a McNemar test~\cite{mcnemar1947note} to determine the statistical significance of differences between our ArtScore and other metrics in terms of their consistency with human artistic evaluation. In this comparison, we exclude ArtFID, as it is based on the deep feature distribution of generated and real artworks and cannot handle instance-level comparison when only limited images are provided. 

The results, shown in Table~\ref{table:evaluation result}, indicate that ArtScore is more \textit{consistent} with human judgment than other metrics when comparing artworks generated with style transfer methods. Notably, ArtScore performs significantly better than Gram Loss, Content Loss, and SSIM when the compared images are based on different style-content references. 
This is in line with our findings in the algorithm-level experiments. Other metrics cannot be directly used to compare the artness of different images, and could be inaccurate in the task of artistic evaluation. Content Loss, SSIM, LPIPS, and L2 measure content preservation based on deep feature or pixel values. Although they could to some extent measure how structural the composition of the generated image is by comparing it with real photo references, they completely \textbf{ignore the style information}. Therefore, they were not able to capture this important aspect for artistic evaluation, such as how the brushstrokes resemble fine-art. Gram Loss, on the other hand, only measures the stylistic and completely \textbf{ignores the contents}. For example, it could ignore artifacts that distort the content, which would otherwise detract from the fine-art feel of the piece. Moreover, when comparing unconditionally generated images, all other metrics would not work as they cannot be derived without paired reference. ArtScore, on the other hand, \textit{remains consistent} with human judgments in this scenario.

For each metric, the accuracy score drops when comparing images with different style-content references. We hypothesize that this is because the task of artistic judgment becomes more difficult and subjective when the compared images are less similar and consistent, even for humans: the ratio of the controversial pairs to all pairs increases from $8.47\%$ to $12.03\%$ when the references for compared images become different.

Interestingly, we notice the performance of Gram Loss is no better than random guess, which again suggests that Gram Loss may not be well aligned with human artistic evaluation.

\vsBfSub
\subsection{Qualitative Analysis}
We conduct a qualitative analysis to gain insights into the differences between generated images with high and low ArtScore. The results are presented in Fig.~\ref{fig:qualitative}. We use images generated in the quantitative experiments and sample examples with varied ArtScore. In the examples generated with NST, we observe that images with high ArtScore tended to have a clearer and more natural structure (\textit{e.g.,} the roof in (a), the people in (b), (c), and the truck in (f)) with fewer introduced artifacts (as seen (d) and (f)). In the examples generated with StyleGAN3, we observe that high-score images tended to have more details ((i) and (j)), a harmonious structure ((h) and (j)), and delicate brushstrokes ((g) and (k)). As a result, high-score images more closely resembled fine-art paintings. These observations demonstrate that the proposed ArtScore effectively captures the qualities of AI-generated images associated with fine art, such as clarity, naturalness, and harmony.

\begin{table}[t]
\centering

\caption{The ablation experiment on instance-level evaluation. We show the performance of the artness evaluator when 1) using only a single domain of images during training, 2) replacing the learn-to-rank objective with the mean squared error loss during training, and 3) removing random swap argumentation during training. The best performance is highlighted in bold, and the second best is underlined.}
\vspace{-0.1in}
\begin{tabularx}{0.48 \textwidth}{p{2.5cm}|>{\centering\arraybackslash}X>{\centering\arraybackslash}X>{\centering\arraybackslash}X}
\hline
\multicolumn{1}{c|}{\multirow{3}{*}{\textbf{Metrics}}} & \multicolumn{3}{c}{\textbf{Accuracy}} \\ \cline{2-4} 
\multicolumn{1}{c|}{} & \begin{tabular}[c]{@{}c@{}}\textbf{NST}\\ \textit{same ref}\end{tabular} & \begin{tabular}[c]{@{}c@{}}\textbf{NST}\\ \textit{diff ref}\end{tabular} & \begin{tabular}[c]{@{}c@{}}\textbf{StyleGAN3}\\ \textit{no ref}\end{tabular} \\ \hline
ArtScore & \textbf{0.6433} & \textbf{0.5639} & \textbf{0.6426} \\
only Face & 0.5205 & 0.5188 & 0.4729 \\
only Horse & 0.5468 & 0.5150 & 0.5054 \\
only Landscape & 0.5877 & 0.5564 & 0.5884 \\
MSE Loss & \underline{0.6404} & \underline{0.5627} & 0.5812 \\
w/o random swap & 0.5556 & 0.5414 & \underline{0.6245} \\ \hline
\end{tabularx}
     \vspace{-0.1in}
\label{table:ablation eval}
\end{table}

\begin{table}[t]
\centering
\caption{The ablation experiment on algorithm-level evaluation. We show the performance of ArtScore when used alone (above) and when used with ArtFID\_Style and Content Loss metrics (below). The best performance is highlighted in bold, and the second best is underlined.}
\vspace{-0.1in}
\begin{tabularx}{0.48 \textwidth}{>{\centering\arraybackslash}Xc|>{\centering\arraybackslash}X>{\centering\arraybackslash}X}
\hline
\multicolumn{2}{c|}{\textbf{Metrics}} & \textbf{Spearman’s \bm{$\rho$} ↑} & \textbf{\textit{p}-value ↓} \\ \hline
\multicolumn{1}{l}{ArtScore} & \multicolumn{1}{c|}{\multirow{6}{*}{}} & \textbf{0.6364} & \textbf{0.0261} \\
\multicolumn{1}{l}{only Face} &  & 0.4336 & 0.1591 \\
\multicolumn{1}{l}{only Horse} &  & 0.5455 & 0.0666 \\
\multicolumn{1}{l}{only Landscape} &  & 0.5594 & 0.0586 \\
\multicolumn{1}{l}{MSE Loss} &  & 0.5594 & 0.0586 \\
\multicolumn{1}{l}{w/o random swap} &  & \underline{0.5734} & \underline{0.0513} \\ \hline
\multicolumn{1}{l}{ArtScore} & \multirow{6}{*}{\begin{tabular}[c]{@{}l@{}}+ArtFID\_Style\\ +Content Loss\end{tabular}} & \textbf{0.8196} & \textbf{0.0011} \\
\multicolumn{1}{l}{only Face} &  & 0.7860 & 0.0024 \\
\multicolumn{1}{l}{only Horse} &  & 0.7951 & 0.0020 \\
\multicolumn{1}{l}{only Landscape} &  & 0.8000 & 0.0018 \\
\multicolumn{1}{l}{MSE Loss} &  & 0.8000 & 0.0018 \\
\multicolumn{1}{l}{w/o random swap} &  & \underline{0.8163} & \underline{0.0012} \\ \hline
\end{tabularx}
  \vspace{-0.25in}
\label{table:ablation corr}
\end{table}

\begin{table}[t]
    \centering
    \caption{Ablation study on algorithm-level evaluation: examining the influence of sequence length and dataset size. We present the performance of ArtScore trained with two variations: 1) decreased sequence length, representing coarser-level interpolation between art and real styles (upper), and 2) subset of the developed dataset, representing fewer sequences during training (bottom). The right-most column indicates the setup we used in the main result.}
    \vspace{-0.1in}
\label{table:added exp}
\begin{tabularx}{0.48 \textwidth}{X|c|c|c|c|c}
\hline
  \textbf{Seq Length} & \textbf{4} &	\textbf{6} & \textbf{8} &	\textbf{10} & \textbf{12} \\ \hline
  \textbf{MSE Loss} & 0.4056 & 0.4406 & 0.5245 & 0.5385 & 0.5594 
\\
  \textbf{ListMLE Loss} & 0.5175 & 	0.5734 & 	0.6084 & 	0.6364 & 	0.6364 \\
\hline
\noalign{\vskip 5pt}
\hline
\textbf{Train Set Ratio} &	\textbf{20\%} &	\textbf{40\%}	& \textbf{60\%} &	\textbf{80\%} &	\textbf{100\%} \\ \hline
\textbf{MSE Loss} & 0.3916 	& 0.5035 	& 0.5175 	& 0.5385 	& 0.5594 \\
\textbf{ListMLE Loss} & 0.4266 	& 0.5804 	& 0.6084 	& 0.6084 	& 0.6364 \\\hline

\end{tabularx}
\vspace{-0.25in}
\end{table}

\vsBfSub
\subsection{Ablation}
The ablation experiment results are reported in Tables~\ref{table:ablation eval}, \ref{table:ablation corr}, and~\ref{table:added exp}. When evaluating the effectiveness of combining ArtScore with other metrics, we adopt the ArtFID\_Style and Content Loss as additional metrics and adopt ranking-based aggregation, as this combination yields the highest correlation as shown in Table~\ref{table:correlation result}. 

\noindent\textbf{Using only a single domain. } We evaluate the performance of ArtScore by training it solely on image sequences belonging to the face, horse, or landscape domain. Compared to using the full dataset, this setting results in a significant drop in performance, highlighting the importance of training the artness evaluator on multiple domains of artwork. This allows the model to learn from a diverse dataset with varying contents and styles. Interestingly, we observe that the face domain yielded the worst result, followed by the horse domain, while the landscape domain resulted in the best performance. We hypothesize that this may be because of the levels of alignment in each domain. The face domain is highly aligned, given that both the source and transferred generators are trained on preprocessed and aligned data. The horse domain is moderately aligned, as it does not require alignment in the training data as long as there is at least one horse in the image. In contrast, the landscape domain is the least aligned, unlike the object constraint in the horse domain or the position constraint in the face domain. Therefore, we hypothesize that a mixed requirement for alignment is more natural in the context of real paintings, as painters may not always set such constraints when creating their works. Overall, combining these domains yields the best results.

\vspace{-0.02in}
\noindent \textbf{Using a Coarser Level of Interpolation. } In the main result, we vary the interpolation parameter $\alpha$ from $0.0$ to $1.0$ in increments of $0.1$, resulting in $12$ images per sequence, including one real image and $11$ interpolated images with varying degrees of artness. We explore variants of pseudo-annotated datasets with fewer images per sequence by varying $\alpha$, which represents coarser differences between art and real styles. As shown in Table~\ref{table:added exp}, longer sequence lengths generally benefit correlation performance. This aligns with our motivation to train ArtScore to distinguish fine-grained differences in the developed dataset, thereby enhancing its robustness during inference.

\vspace{-0.02in}
\noindent \textbf{Using Fewer Sequences. } We train ArtScore with only a subset of the developed dataset. As depicted in Table~\ref{table:added exp}, the model benefits when trained with more sequences. A more comprehensive and diverse training set enhances the model's generalizability and improves performance on unseen images during inference.

\vspace{-0.02in}
\noindent\textbf{Replacing the learn-to-rank objective. } We replace the ListMLE loss with the MSE loss to train ArtScore. The model is then optimized to output an accurate value of the pseudo ranking for the input image sequences. Our evaluation results (Table~\ref{table:ablation eval}, \ref{table:ablation corr}) demonstrate that ListMLE outperforms MSE in both instance- and algorithm-level evaluations, across various conditions such as sequence length and training set size (Table~\ref{table:added exp}). With the MSE loss, the model is encouraged to give the same scores for images with the same ranking in their respective sequences, even though their ground-truth artness could vary. This additional constraint could hamper the learning process.

\vspace{-0.02in}
\noindent\textbf{Removing random swap argumentation. } We remove the data augmentation of random swap between different sequences when training the artness evaluator. Similarly, we observe a drop in performance. This experiment shows the effectiveness of random swap augmentation, which can provide a scenario where ArtScore needs to compare the artness between images with different contents and styles, making it more robust during inference.

\vsBf
\section{Conclusion and Future Work}
In this paper, we present a novel metric named ArtScore for evaluating the quality of AI-generated images based on their similarity to fine-art paintings. Our framework, which includes StyleGAN adaptation, interpolation image generation, and ArtScore training, allows for the automatic labeling of artness levels, thus enabling fair and quantitative comparisons between different algorithms. Our experiments demonstrate that the proposed ArtScore aligns better with human artistic judgment than all existing metrics. 
Furthermore, combining ArtScore with selected existing metrics provides a highly accurate assessment of the {\it overall quality} of the generated images. Used alone or in combination, ArtScore provides a valuable tool for researchers and artists in the field of image generation. 
We hope that our work will contribute to the advancement of the field by inspiring more research in developing reliable metrics for artistic evaluation. 

In future work, we aim to further develop ArtScore by incorporating it into the training or generation process of image generative models as a loss term. For example, ArtScore can be used as a guidance function~\cite{bansal2023universal} for diffusion model sampling and to regularize diffusion model training~\cite{black2023training}, leading to results more closely resembling fine-art paintings. While our focus primarily aligns with AI researchers, we acknowledge that artists and art researchers might have more nuanced perspectives on the concept of ``artness." Currently, no dataset provides such annotations, and gathering data from experts presents challenges due to cost, diversity, and the subjectivity inherent in artistic judgment. Consequently, we used automatic methods and user studies to evaluate artness, following common practices in related research. In future work, we aim to integrate insights from experts through case studies to enrich our understanding of artness.

\vsBf
\section*{Acknowledgments}
\noindent This work is supported in part by the Goergen Institute for Data Science at the University of Rochester. 

\vsBf
{
\small
\bibliographystyle{IEEEtran}
\bibliography{report}
}

\begin{IEEEbiography}[{\includegraphics[width=1in,height=1.25in,clip,keepaspectratio]{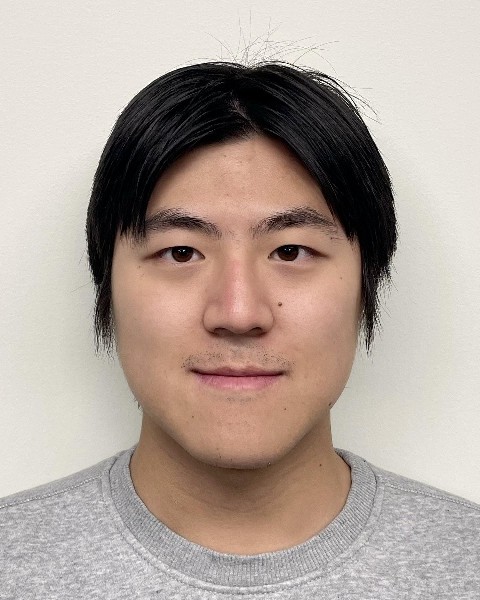}}]{Junyu Chen}
received the B.S. degree in psychology and the B.A degree in Economics from Peking University, Beijing, China, in 2020, and the M.S. degree in data science from the University of Rochester, Rochester, NY, USA, in 2022, where he is currently pursuing the Ph.D. degree with the Department of Computer Science. \\
His general research interests are computer vision,  continual learning, and deep generative model.
\end{IEEEbiography}

\vspace{11pt}
\vspace{-33pt}
\begin{IEEEbiography}[{\includegraphics[width=1in,height=1.25in,clip,keepaspectratio]{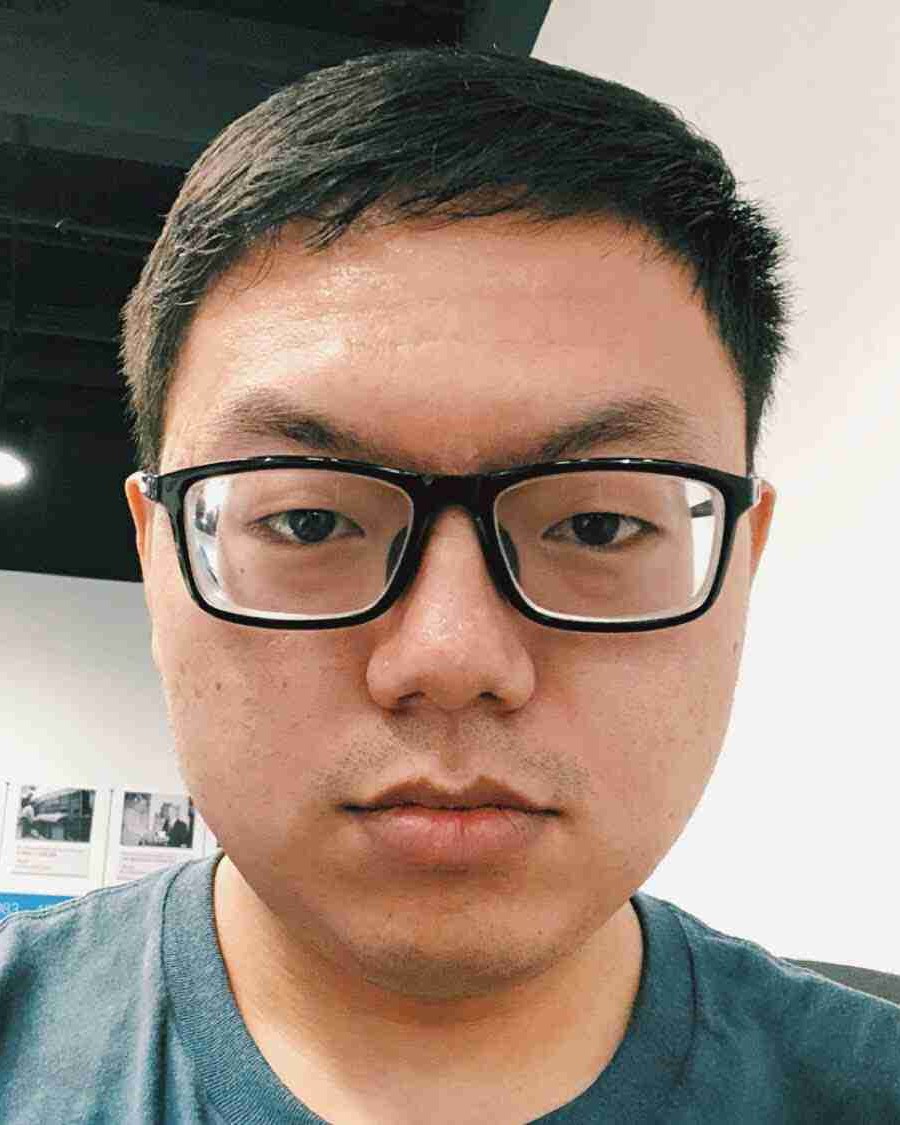}}]{Jie An}
received the B.S. degree in information and computing sciences and M.S. degree in applied mathematics from Peking University, Beijing, China, in 2016 and 2019, respectively. He is currently pursuing the Ph.D. degree with the Department of Computer Science, University of Rochester, Rochester, NY, USA.\\
His general research interests are computer vision, deep generative model, and AI+Art.
\end{IEEEbiography}

\vspace{11pt}
\vspace{-33pt}
\begin{IEEEbiography}[{\includegraphics[width=1in,height=1.25in,clip,keepaspectratio]{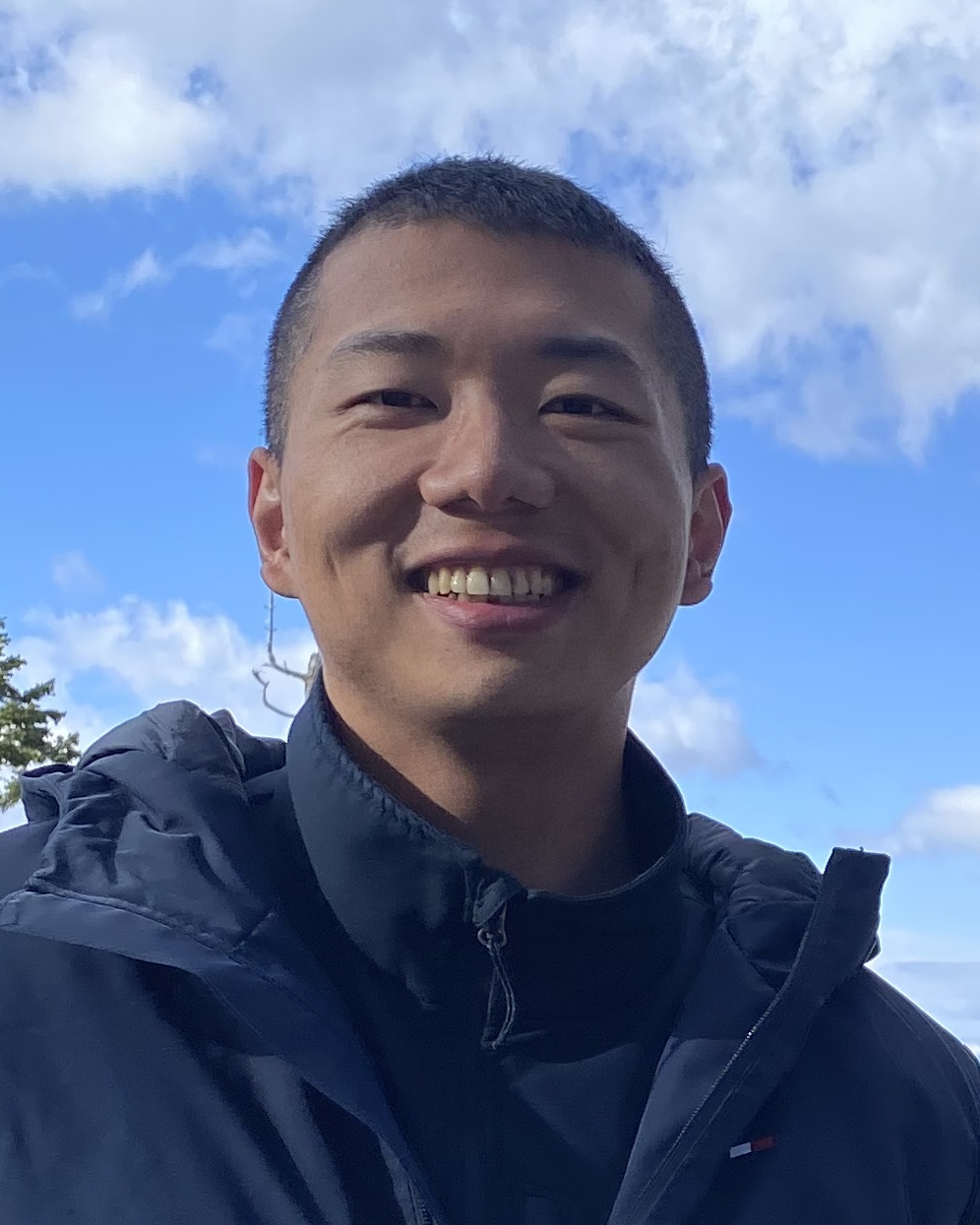}}]{Hanjia Lyu}
received the B.A. degree in economics from Fudan University, Shanghai, China, in 2017, and the M.S. degree in data science from the University of Rochester, Rochester, NY, USA, in 2020, where he is currently pursuing the Ph.D. degree with the Department of Computer Science. \\
His general research interests are data mining, network science, computational social science, machine learning, and health informatics.
\end{IEEEbiography}

\vspace{11pt}
\vspace{-33pt}
\begin{IEEEbiography}[{\includegraphics[width=1in,height=1.25in,clip,keepaspectratio]{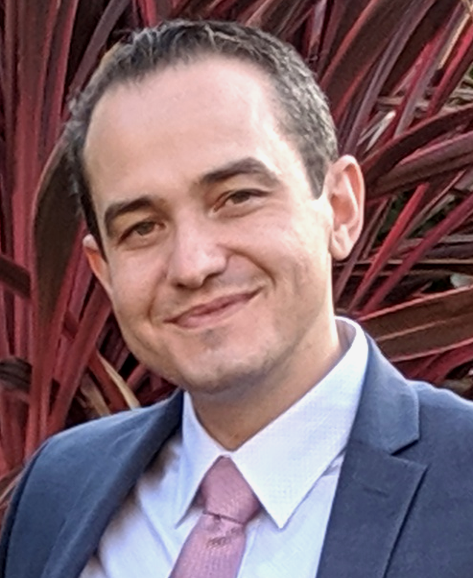}}]{Christopher Kanan}
(Senior Member, IEEE) received the B.S. degree in computer science and philosophy from Oklahoma State University, Stillwater, OK, USA, in 2004, the M.S. degree in computer science from the University of Southern California, Los Angeles, CA, USA, in 2006, and the Ph.D. degree from the University of California at San Diego, San Diego, CA, in 2013, where he worked on machine learning, cognitive science, and computer vision. After completing his Ph.D., he was a Post-Doctoral Scholar at Caltech, Pasadena, CA, and then was a Research Technologist with the NASA Jet Propulsion Laboratory until 2015. He is currently an Associate Professor of Computer Science at the University of Rochester, Rochester, NY, USA. His lab’s main focus is basic research in deep learning, with an emphasis on lifelong (continual) machine learning, bias-robust artificial intelligence, medical computer vision, and language-guided scene understanding.
\end{IEEEbiography}

\begin{IEEEbiography}[{\includegraphics[width=1in,height=1.25in,clip,keepaspectratio]{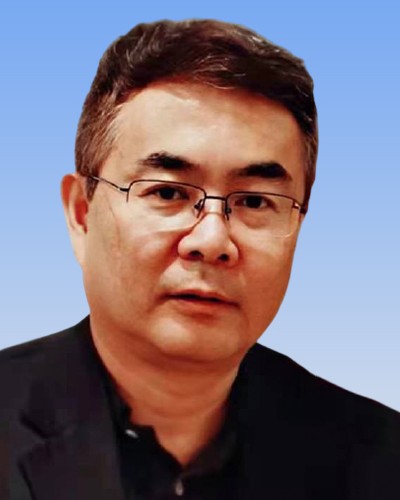}}]{Jiebo Luo}
(Fellow, IEEE) joined the Department of Computer Science, University of Rochester, Rochester, NY, USA, in 2011, after a prolific career of over 15 years with Kodak Research Laboratories, Rochester. He has authored over 500 technical articles and holds over 90 U.S. patents. His research interests include computer vision, natural language processing, machine learning, data mining, computational social science, and digital health. \\
Prof. Luo is also a fellow of the Association for Computing Machinery (ACM), the Association for the Advancement of Artificial Intelligence (AAAI), the International Society for Optics and Photonics (SPIE), and the International Association for Pattern Recognition (IAPR). He is the Editor-in-Chief of the IEEE TRANSACTIONS ON MULTIMEDIA for the term 2020–2022. He has served as the Program Co-Chair for the 2010 ACM Multimedia, 2012 IEEE Conference on Computer Vision and Pattern Recognition (CVPR), 2016 ACM International Conference on Multimedia Retrieval (ICMR), and 2017 IEEE International Conference on Image Processing (ICIP) and the General Co-Chair for the 2018 ACM Multimedia Conference. He has served on the Editorial Board of the \textsc{IEEE Transactions on Pattern Analysis and Machine Intelligence}, \textsc{IEEE Transactions on Multimedia}, \textsc{IEEE Transactions on Circuits and Systems for Video Technology}, \textsc{IEEE Transactions on Big Data}, \textit{Pattern Recognition}, \textit{Machine Vision and Applications}, and \textit{ACM Transactions on Intelligent Systems and Technology}.
\end{IEEEbiography}

\end{document}